\documentclass{article}

\PassOptionsToPackage{numbers, compress}{natbib}

\usepackage[final]{AdvML_Frontiers_2024}

\usepackage[utf8]{inputenc} 
\usepackage[T1]{fontenc}    
\usepackage{url}            
\usepackage{booktabs}       
\usepackage{amsfonts}       
\usepackage{nicefrac}       
\usepackage{microtype}      
\usepackage{xcolor}         

\usepackage{wrapfig}
\usepackage{graphicx}
\usepackage{subcaption}
\usepackage{amsmath}
\usepackage{multirow}
\usepackage{makecell}
\usepackage{colortbl}
\usepackage{wrapfig} 
\usepackage{fancyhdr}

\title{Hiding-in-Plain-Sight (HiPS) Attack on CLIP for Targetted Object Removal from Images}

\author{%
Arka Daw$^{1}\footnotemark[1]$ \quad Megan Hong-Thanh Chung$^{1}\thanks{Equal Contribution}$ \quad Maria Mahbub$^1$ \quad Amir Sadovnik$^1$ \\
$^1$ Oak Ridge National Laboratory (ORNL)\\
}

\begin{document}

\maketitle

\vspace{-3ex}
\begin{abstract}
\vspace{-2ex}
Machine learning models are known to be vulnerable to adversarial attacks, but traditional attacks have mostly focused on single-modalities. With the rise of large multi-modal models (LMMs) like CLIP, which combine vision and language capabilities, new vulnerabilities have emerged. However, prior work in multimodal targeted attacks aim to completely change the model's output to what the adversary wants. In many realistic scenarios, an adversary might seek to make \emph{only subtle modifications to the output}, so that the changes go unnoticed by downstream models or even by humans. We introduce \emph{Hiding-in-Plain-Sight (HiPS)} attacks, a novel class of adversarial attacks that subtly modifies model predictions by selectively concealing target object(s), as if the target object was absent from the scene. We propose two HiPS attack variants, HiPS-cls and HiPS-cap, and demonstrate their effectiveness in transferring to downstream image captioning models, such as CLIP-Cap, for targeted object removal from image captions. 
\end{abstract} 

\vspace{-3ex}
\section{Introduction}
The vulnerability of machine learning (ML) models to adversarial attacks—small perturbations in input data that lead to incorrect predictions—has been extensively studied \citep{goodfellow2014explaining, madry2017towards} across various domains, including image classification \citep{machado2021adversarial, chakraborty2018adversarial} and biometrics \citep{brown2017adversarial, jia2022adv}. However, most existing adversarial attacks have been designed for single modalities, primarily focusing on the image or, less frequently, the text domain \citep{jia2017adversarial, ebrahimi2017hotflip}. The advent of large foundational models, such as large language models (LLMs) \citep{zhao2023survey} and large multi-modal models (LMMs) \citep{yang2023dawn} (e.g., Chat-GPT \citep{achiam2023gpt}, Gemini \citep{team2023gemini}), which have shown great promise across a diverse range of tasks \citep{hendrycks2020measuring} (such as zero-shot classification, visual question answering, and image captioning) has revolutionized the ML community and led to their widespread adoption. Many of these models \citep{liu2024visual} integrate a pre-trained LLM with a large vision encoder, such as CLIP \citep{radford2021learning} which is a foundational multimodal model trained on 400M image-text pairs via contrastive learning. Typically, the vision encoder of such LMMs remains frozen during training, and the vision embeddings are mapped into the shared embedding space of the LLM using a simple projection layer. However, this introduces an obvious vulnerability \citep{shayegani2023plug}: adversarial attacks developed against these open-source vision encoders (such as CLIP) can be directly transferred to LMMs, compromising their integrity.

While generating adversarial attacks on LMMs (such as CLIP) \citep{shayegani2023plug, schlarmann2024robust} has been explored, generally termed as \emph{jailbreaking LMMs} \citep{guo2024cold, zhou2024easyjailbreak}, they have primarily focused on `completely' changing the output of the LMM to what the adversary wants, which is typically very different from the original outputs (without any perturbation). However, in many real-world scenarios, an adversary might seek to make only `subtle' modifications to the output, so that the changes go unnoticed by downstream models or even by humans. To this end, we introduce a novel class of adversarial attacks on images, termed \emph{Hiding-in-Plain-Sight (HiPS)} attacks. The primary goal of a HiPS attack is to generate an adversarial image that subtly modifies the model’s predictions by selectively concealing a specific target object while leaving the rest of the model's functionality intact. For example, a HiPS adversarial image designed to hide a particular object should cause an image captioning model to generate a caption as if the target object(s) was never present, while the rest of the image content should stay in tact. We propose two distinct types of HiPS attacks using the CLIP vision encoder: (1) HiPS-cls, which generates the attack by leveraging only the class label information, and (2) HiPS-cap, which utilizes the original image caption and a target caption to craft the attack. We demonstrate that our HiPS attacks can effectively transfer to downstream image captioning models, such as CLIP-Cap \citep{mokady2021clipcap}, enabling selective removal of target objects from image captions. Additionally, we introduce several novel evaluation metrics to assess the performance of our proposed HiPS attacks in targeted object removal.

\section{Background and Related Works}
\label{sec:background}
\textbf{Adversarial Robustness:} 
One of the seminal methods for generating adversarial attacks is the Fast Gradient Sign Method (FGSM) \citep{goodfellow2014explaining}, a simple, single-step $L_\infty$-bounded attack, defined as: $I_{\text{adv}} = I + \epsilon \text{sign}(\nabla_I \mathcal{L}(I, y))$,
where $I_{\text{adv}}$ is the adversarial image, $I$ is the original image, $\epsilon$ is the attack budget, and $\mathcal{L}$ is the loss function to be maximized for the attack. For an untargeted attack, $\mathcal{L}$ is typically the cross-entropy loss with respect to the correct class $y$, 
and in a targeted attack, the objective shifts to minimizing the loss with respect to a target class $\tilde{y}$, making $\mathcal{L}$ the negative cross-entropy loss for the target class. Another widely used technique is the Projected Gradient Descent (PGD) attack \citep{madry2017towards}, which is the strongest first-order attack. PGD is an iterative, first-order optimization-based attack, defined as: $I^{t+1} = \mathcal{P}_{I+\mathcal{S}}(I^t + \alpha \text{sign}(\nabla_I \mathcal{L}(I, y)))$,
where $t$ denotes the iteration number, $\mathcal{P}$ is a projection operation that maps the perturbed input back onto a $L_p$ ball with radius $\epsilon$, with $\mathcal{S}$ representing the region defined by the $L_p$ ball, and $\alpha$ is the step size. 

\textbf{Multi-modal Models:} CLIP \citep{radford2021learning} is one of the seminal works in multi-modal modeling 
due to its exceptional performance in zero-shot tasks.
Recently, there has been growing popularity in developing large multi-modal models (LMMs) \citep{yang2023dawn} (GPT-4V \citep{achiam2023gpt}, Gemini \citep{team2023gemini}, LLaVA \citep{liu2024visual})
driven by their impressive capabilities across a wide range of tasks and domains \citep{hendrycks2020measuring}. Many of these models integrate a pre-trained large language model (LLM), such as Llama \citep{touvron2023llama} or Vicuna \citep{peng2023instruction}, with a large vision encoder like CLIP. For LLaVA, the vision encoder remains frozen during training, with a simple projection layer mapping the vision embeddings to the shared embedding space of the LLM. 

\textbf{Adversarial Robustness of LMMs:} With the advent of LMMs, investigating their vulnerabilities has become an important research focus in AdvML, often referred to as \emph{jailbreaking} LLMs and LMMs \citep{zhou2024easyjailbreak, guo2024cold}. While previous studies have demonstrated that jailbreaking LLMs is feasible with full access to model parameters, recent findings highlight that LMMs are particularly susceptible to adversarial attacks targeting the vision modality \citep{shayegani2023jailbreak}. In particular, even with access solely to the vision encoder, such as the open-sourced CLIP model--adversaries can exploit these vulnerabilities to jailbreak LMMs like LLaVA and OpenFlamingo\citep{awadalla2023openflamingo}, which rely on the frozen CLIP vision encoder.

\section{Hiding-in-Plain-Sight (HiPS) Attack}

\begin{figure}[t]
\centering
    \includegraphics[width=\textwidth]{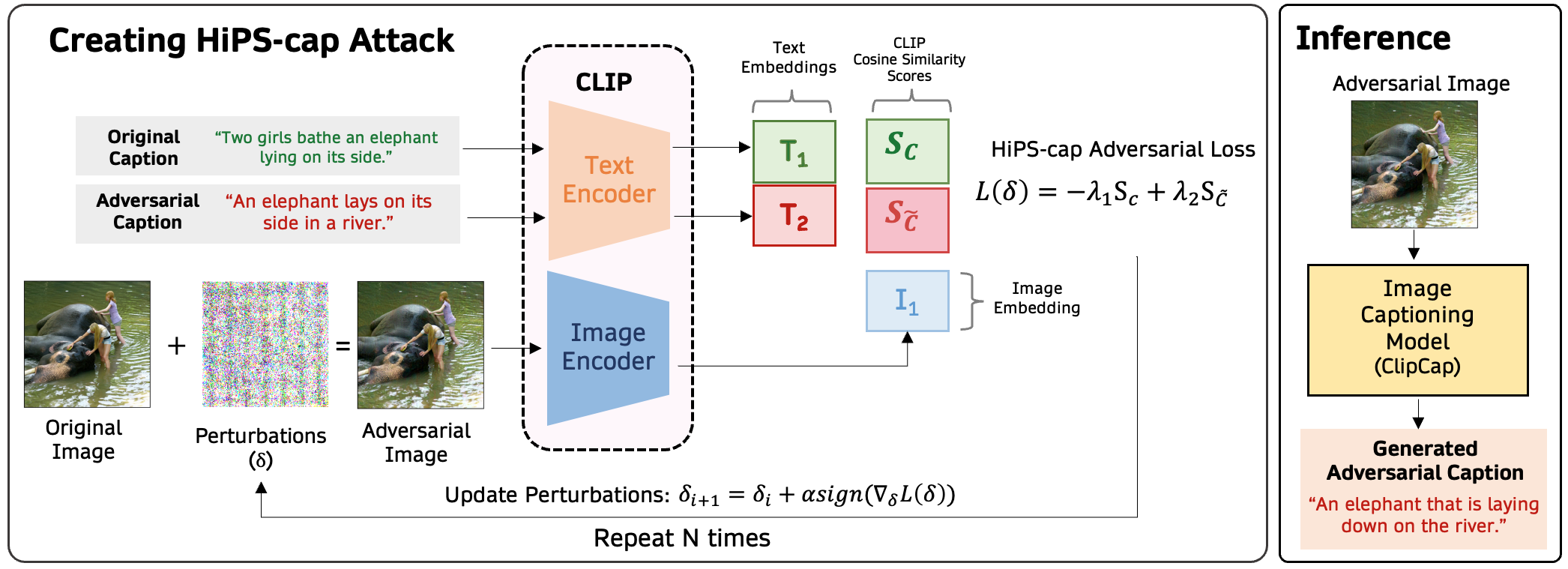}
    \caption{A schematic illustration of the \emph{Hiding-in-Plain-Sight} (HiPS-cap) Attack.}
    \label{fig:clip_attack}
\end{figure}

Traditional `targeted' adversarial attacks on images are designed to drastically alter the behavior of a downstream ML models (such as a Large Vision Language Model or an image classifier), forcing them to produce outputs that align with the adversary's objectives. In contrast, we introduce a novel class of adversarial attacks on images, termed \textit{Hiding-in-Plain-Sight (HiPS)} attacks. The primary goal of a HiPS attack is to generate an adversarial image that can \textit{`subtly'} modify the model(s) predictions by selectively concealing a specific `target' object while leaving the rest of the model's functionality intact. For instance, a HiPS adversarial image designed to conceal (or `target') a particular object should cause an image captioning model to generate a caption as if the target object(s) was never present while the rest of the image content should stay in tact. 
Similarly, when a HiPS adversarial image is processed by a LMM, the model should respond to queries about the image as if the target object were absent. Ideally, the adversarial images generated using the HiPS attacks should be universal and transferable across a variety of downstream ML models. Therefore, generating the HiPS attack using a \emph{`foundation'} multi-modal model that is already universally used for a variety of downstream tasks is necessary for transferability. 
For simplicity, in this paper, we focus on investigating the transferability of HiPS attacks specifically on image captioning models \citep{stefanini2022show}.





\vspace{-2ex}
\subsection{Problem Formulation}
In this section, we introduce the formal notations used throughout this paper. Let $I$ represent an input image containing $n$ different object classes, and let $\mathcal{T}_{I} = \{ T_1, T_2, \dots, T_n \}$ denote the set of objects present in the image $I$, where $T_i$ corresponds to the textual description (or simply, the class labels) of the $i$-th object. The target object to be removed is denoted as $T_{\text{target}} = T_j$ for some $j \in \{1, 2, \dots, n\}$. We will utilize the CLIP model to generate the HiPS attack, which consists of an image encoder, $f_{\mathtt{Image}}: I \rightarrow \mathbf{Z}_{\mathtt{Image}}$, and a text encoder, $f_{\mathtt{Text}}: T \rightarrow \mathbf{Z}_{\mathtt{Text}}$, where $T$ is a textual input, $\mathbf{Z}_{\mathtt{Image}} \in \mathbb{R}^D$ is the image embedding, $\mathbf{Z}_{\mathtt{Text}} \in \mathbb{R}^D$ is the text embedding, and $D$ is the embedding dimension.

In the context of image captioning, the objective of the HiPS attack is to generate an adversarial image $I_{\text{adv}}$ that is nearly indistinguishable from $I$. However, when this adversarial image $I_{\text{adv}}$ is processed by the downstream image captioning model, $f_{\text{caption}}: I \rightarrow T$, the generated caption $\hat{C}_{\text{adv}}=f_{\text{caption}}(I_{\text{adv}})$ should omit the target object $T_{\text{target}}$ while accurately describing all other objects in the image.  In other words, the adversarially generated caption, $\hat{C}_{\text{adv}}$, should \emph{closely resemble} the caption $\hat{C}_{\text{orig}}$ produced from the original (unperturbed) image, with the exception that $T_{\text{target}}$ is not mentioned. 
This approach contrasts sharply with traditional targeted attacks, where typically the goal is to produce an output that is significantly different from the correct one. 
While a traditional adversarial attack's goal is to make the perturbation imperceptible in the \emph{input space}, in the HiPS attack, we want the difference in the output space to also be minimal - the only difference should be the omission of the target class.
In this paper, we propose two different variants of the HiPS attack in the context of image captioning, which are detailed below (See Figure \ref{fig:clip_attack} for a schematic representation of HiPS Attack).

\subsection{HiPS-cls Attack using Class Labels}
In this variant of HiPS attack, termed \emph{HiPS-cls}, we utilize only the textual class labels $\mathcal{T}_I$ to obtain the adversarial image. Given an image $I$ and its corresponding set of class labels $\mathcal{T}_I$, we compute the cosine similarity scores $S_i$ for each class label $T_i$ as follows:
\begin{equation}
    S_i = \cos(f_{\mathtt{Image}}(I), f_{\mathtt{Text}}(T_i)) = \left \langle \frac{f_{\mathtt{Image}}(I)}{\lVert f_{\mathtt{Image}}(I) \rVert_2},  \frac{f_{\mathtt{Text}}(T_i)}{\lVert f_{\mathtt{Text}}(T_i) \rVert_2} \right\rangle
\end{equation}

The cosine similarity $S_i$ between the image $I$ and class label $T_i$ measures the alignment between their respective image and text embeddings. A higher score $S_i$ indicates that the object with class label $T_i$ is likely present in the image $I$, while a lower score suggests its absence. Since the objective of the HiPS attack is to remove the target object $T_{\text{target}} = T_j$, our goal is to perturb the image $I$ in such a way that the cosine similarity score for the target object, $S_j$, is reduced (as if it is absent), while the scores for all other objects $T_i$ (for all $i \neq j$) are either increased or remain unchanged.
To formalize this, we define the HiPS-cls adversarial loss function as follows: 
$\mathcal{L}_{\text{HiPS-cls}} = -\lambda_1 S_j + \lambda_2 \sum_{i \neq j} S_i$.

\subsection{HiPS-cap Attack using Adversarial Captions}
In this variant of the HiPS attack, termed \emph{HiPS-cap}, rather than using class labels, we generate the attack on CLIP by utilizing the original caption $C$ and a target caption $\tilde{C}$. The target caption $\tilde{C}$ is designed to be similar to $C$, but as if the target object $T_{\text{target}}$ were not present in the image. In other words, $\tilde{C}$ represents an ideal adversarial caption that a successful HiPS attack on a captioning model should produce.
Similar to the HiPS-cls approach, we calculate the cosine similarities between the image $I$ and both the original caption $C$ and the target caption $\tilde{C}$ as follows:
\begin{equation}
    S_C = \cos(f_{\mathtt{Image}}(I), f_{\mathtt{Text}}(C)); \quad\quad S_{\tilde{C}} = \cos(f_{\mathtt{Image}}(I), f_{\mathtt{Text}}(\tilde{C}))
\end{equation}
The corresponding adversarial loss can be computed as $\mathcal{L}_{\text{HiPS-cap}} = -\lambda_1 S_C + \lambda_2 S_{\tilde{C}}$. $\mathcal{L}_{\text{HiPS-cap}}$ aims to reduce the score for the original caption $S_C$ while increase the score for the target caption $S_{\tilde{C}}$, where the target object is missing.
The adversarial loss $\mathcal{L}_{\text{HiPS-cls}}$ and $\mathcal{L}_{\text{HiPS-cap}}$ can be optimized using existing adversarial attacks such as FGSM and PGD attacks (See Section \ref{sec:background}).



\section{Experimental Setup}

\textbf{Setting:} We develop HiPS-cls and HiPS-cap attacks using the CLIP model, where the vision encoder is based on Vision Transformer architecture (ViT-B/32) \citep{dosovitskiy2020image}. To generate adversarial images for the HiPS attack, we employ established techniques, including FGSM and PGD with $L_\infty$, $L_1$, and $L_2$ norms. For simplicity, we focus on images containing only two foreground objects: one serving as the target object to be removed, and the other as the object to be retained in the adversarial caption. We manually sampled 50 such images from the MS COCO dataset to test our two HiPS attack variants ($cap$ vs. $cls$). For the HiPS-cap attack, we use the original COCO captions as $C$, and manually generate two target (adversarial) captions, one used for training ($\tilde{C}$), while the other one is reserved for evaluation. For the downstream captioning model, we utilize the CLIP-Cap \citep{mokady2021clipcap} model. CLIP-Cap uses the vision encoder from CLIP and a mapping network to project the image embeddings into a shared representation space, where a language model (GPT-2) \citep{radford2019language} generates the captions. 


\textbf{Evaluation Metrics:} In the context of assessing the success of HiPS attack, we introduce several novel metric to measure attack success, where we consider two main criterions. First, the ability to successfully remove references of the target object from the generated textual caption. We propose a metric called \emph{Target Object Removal Rate (TORR)} to capture this using similarity-based assessments and string-matching comparisons between words. Second, the ability to measure if the remaining objects are intact to ensure that perturbation does not inadvertently affect or remove references to the objects other than the targeted one. We propose another metric called \emph{Remaining Objects Retention Rate (RORR)} for this purpose. Next, we utilize Attack Success Rate (ASR) that measures if both of these criterions (TORR and RORR) are satisfied. We additionally utilize Caption Semantic Similarity (CSS) which is essentially the cosine similarity between the ground truth adversarial caption, and the generated adversarial caption ($\cos(\tilde{C}_{gt}, \hat{C}_{\text{adv}})$). CSS measures if the two are semantically close to each other in the text embedding space. Additional details of computation of TORR, RORR and ASR are provided in the Appendix. The image quality is another important metric to measure the imperceptibility of the attack. We use standard metrics such as Mean Squared Error (MSE), Mean Absolute Error (MAE), Signal-to-Noise Ratio (PSNR), and Structural Similarity metric (SSIM).

\textbf{Baselines:} We compare against two PGD ($L_\infty$) based attacks: targeted and untargeted. For the class-labels variant, in the PGD (targeted) setting, we set $\lambda_1=1$, $\lambda_2=0$, focusing solely on removing the target object. In the PGD (untargeted) setting, we set $\lambda_1=0$, $\lambda_2=1$, prioritizing the retention of all other objects in the image. For adversarial captions variant, we only use PGD (targetted) setting, where we optimize to maximize the similarity with target caption  (  $\lambda_1=0$, $\lambda_2=1$).


\vspace{-2ex}
\section{Results}
\begin{figure}[ht]
\centering
    \includegraphics[width=1.02\linewidth]{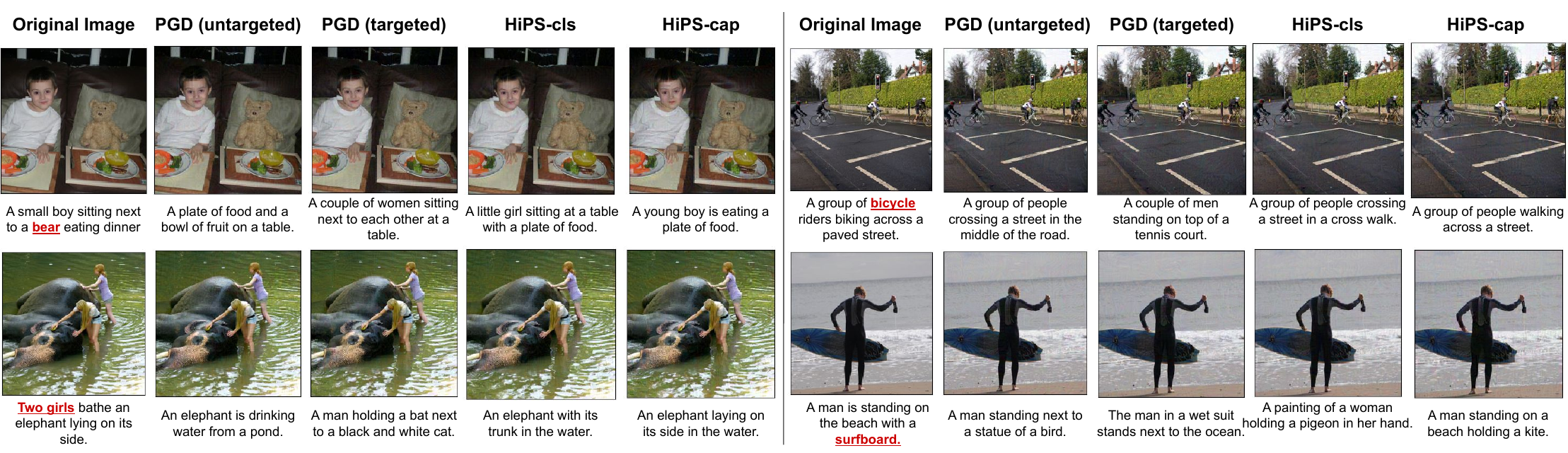}
    \caption{Qualitative Results comparing various methods with \textcolor{red}{target} shown as \textcolor{red}{red} words of caption.}
    \label{fig:qualitative_results}
\end{figure}

\vspace{-2ex}
\textbf{Quantitative Evaluation of HiPS-cls and HiPS-cap:} In Tables \ref{tab:asr} and \ref{tab:image_qual}, we compare the attack success and image quality metrics of the two HiPS variants, using FGSM and PGD under $L_\infty$, $L_1$, and $L_2$ norm constraints. We report results for the best-performing model in each case (see hyper-parameter settings in Appendix). FGSM performs poorly across both HiPS variants, achieving an ASR of only 36-38\%. In contrast, the various PGD attacks demonstrate strong performance across both variants, with the $L_\infty$ variant slightly outperforming the $L_1$ and $L_2$ norms. Specifically, for the HiPS-cls attack, PGD achieves 100\% RORR, indicating that the adversarial captions consistently retain the non-target objects, and TORR of 90\% or higher, demonstrating effective removal of the target object from the caption. However, the TORR, RORR, and ASR metrics for HiPS-cap are slightly lower than those for HiPS-cls (across all PGD attack norms), while the CSS is significantly higher. This indicates that while HiPS-cls is more effective at retaining non-target objects and removing target objects, but it produces lower-quality captions, often resulting in grammatical errors and introducing unnatural artifacts that negatively impact CSS (See Figure \ref{fig:qualitative_results} for a qualitative comparison of different methods). Further, we observe that the baseline method PGD (targeted) achieves a high TORR of 90\% but a relatively low RORR of 78\%, indicating that it primarily optimizes for the removal of the target object. In contrast, PGD (untargeted) exhibits the opposite trend, prioritizing the retention of non-target objects, achieving a 100\% RORR and 76\% TORR. For PGD (targeted) using the adversarial caption, we observe that solely optimizing for similarity to the adversarial caption results in a lower 66\% ASR. This occurs because the adversarial caption is, by definition, already similar to the original caption, making it less effective at removing the target object. To achieve a higher TORR, it is necessary to move away from the original caption (i.e., $\lambda_1 > 0$). Additionally, we compare the image quality metrics in Table \ref{tab:image_qual}. As expected, FGSM utilizes the entire attack budget and introduces significantly larger perturbations compared to the iterative PGD attack. We also observe that the perturbations introduced by PGD under the $L_1$ norm are slightly smaller when compared to $L_2$ and $L_\infty$, as the $L_1$ norm favors localized perturbations with minimal overall change.
\begin{table}[h]
\centering
{\footnotesize
\caption{Comparison of attack success metrics for HiPS-cls and HiPS-cap attacks, optimized using FGSM and PGD under $L_\infty$, $L_1$, and $L_2$ norm constraints.}
\label{tab:asr}
\renewcommand{\arraystretch}{1.1}
\begin{tabular}{ccccccccc}
\hline
\multicolumn{1}{l}{} & \multicolumn{4}{c}{HiPS-cls (Class Labels)}                      & \multicolumn{4}{c}{HiPS-cap (Adv. Caption)}                     \\ \hline
\multicolumn{1}{l}{} & TORR ↑        & RORR ↑         & ASR ↑         & CSS ↑           & TORR ↑        & RORR ↑        & ASR ↑         & CSS ↑           \\ \hline
FGSM                 & 38.0          & 98.0           & 36.0          & 0.6907          & 40.0          & 96.0          & 38.0          & 0.7066          \\
PGD ($L_1$)          & 88.0          & \textbf{100.0} & 88.0          & 0.6898          & 84.0          & \textbf{98.0} & 84.0          & 0.7578          \\
PGD ($L_2$)          & 90.0          & \textbf{100.0} & 90.0          & 0.6701          & 88.0          & 96.0          & 86.0          & 0.7546          \\
PGD ($L_\infty$)     & \textbf{94.0} & \textbf{100.0} & \textbf{94.0} & \textbf{0.6901} & \textbf{90.0} & \textbf{98.0} & \textbf{90.0} & \textbf{0.7673} \\ \hline
PGD (untarget)       & 76.0          & \textbf{100.0} & 76.0          & 0.6790          & \textbf{-}    & \textbf{-}    & \textbf{-}    & \textbf{-}      \\
PGD (target)         & 90.0          & 78.0           & 72.0          & 0.6111          & 66.0          & 100.0         & 66.0          & 0.7499          \\ \hline
\end{tabular}}
\end{table}
\vspace{-4ex}
\begin{table}[ht]
\centering
{\footnotesize
\caption{Comparison of image quality metrics for HiPS-cls and HiPS-cap attacks, optimized using FGSM and PGD under $L_\infty$, $L_1$, and $L_2$ norm constraints.}
\label{tab:image_qual}
\renewcommand{\arraystretch}{1.1}
\begin{tabular}{ccccccccc}
\hline
\multicolumn{1}{l}{} & \multicolumn{4}{c}{HiPS-cls (Class Labels)}                                        & \multicolumn{4}{c}{HiPS-cap (Adv. Caption)}                                \\ \hline
\multicolumn{1}{l}{} & MSE ↓  & MAE ↓ & PSNR ↑ & SSIM ↑  & MSE ↓  & MAE ↓  & PSNR ↑ & SSIM ↑  \\ \hline
FGSM                 & 61.56 & 7.81 & 30.25                                                  & 79.39 & 61.56 & 7.81 & 30.25                                                  & 79.46 \\
PGD ($L_\infty$)           & 12.32 & 3.03 & 37.2                                                   & 94.84 & 23.04 & 3.69 & 34.62                                                  & 92.23 \\
PGD ($L_1$)             & \textbf{6.67}  & \textbf{1.73} & \textbf{39.91}                                                  & \textbf{97.02} & \textbf{6.46}  & \textbf{1.73} & \textbf{40.05}                                                  & \textbf{97.14} \\
PGD ($L_2$)             & 11.33 & 2.23 & 37.60                                                  & 95.32 & 35.36 & 4.18 & 32.67                                                  & 88.93 \\ \hline
\end{tabular}}
\end{table}

\begin{figure}[ht]
\centering
    \includegraphics[width=\linewidth]{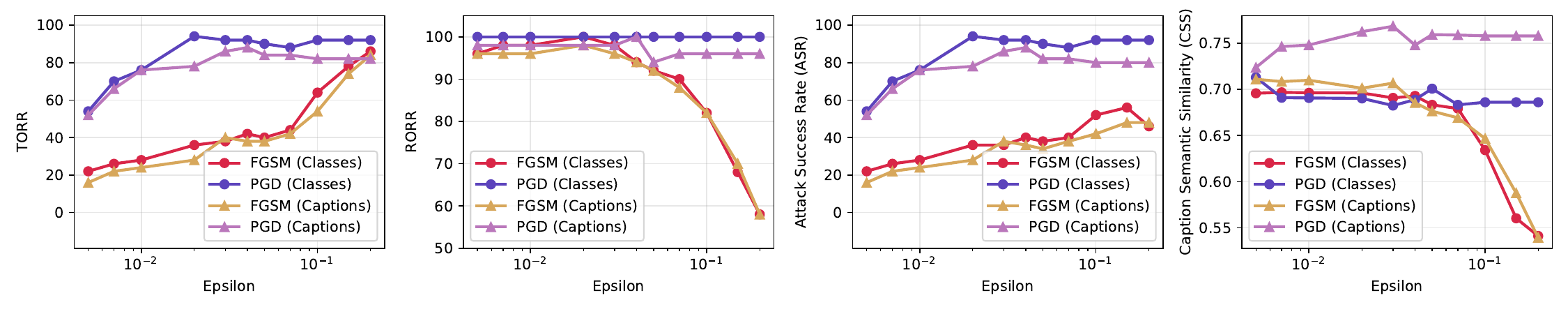}
    \caption{Comparing the effect of attack budget $\epsilon$ on the different attack success metrics for HiPS-cls and HiPS-cap attacks using FGSM and PGD with $L_\infty$ norm.}
    \label{fig:epsilon_sweep_asr}
\end{figure}

\textbf{Effect of Attack Budget:} Figure \ref{fig:epsilon_sweep_asr} illustrates the impact of the attack budget $\epsilon$ on various attack success metrics for HiPS-cls and HiPS-cap attacks using FGSM and PGD with the $L_\infty$ norm. Consistent with previous observations, FGSM performs significantly worse than PGD across all attack budgets for both HiPS variants. For PGD attacks, as expected, increasing the $\epsilon$ value leads to an improvement in the ASR up to $\epsilon=0.05$, after which the ASR gradually saturates, with HiPS-cls showing slightly better performance than HiPS-cap. However, it is noteworthy that as $\epsilon$ increases, the CSS for HiPS-cls drops sharply, whereas HiPS-cap maintains a relatively stable CSS around 0.75. Image quality metrics for the various attack budgets are presented in Appendix Figure \ref{fig:epsilon_sweep_img_qual}. 

\begin{figure}[h]
    \centering
    \begin{subfigure}[b]{0.49\textwidth}
        \centering
        \includegraphics[width=\textwidth]{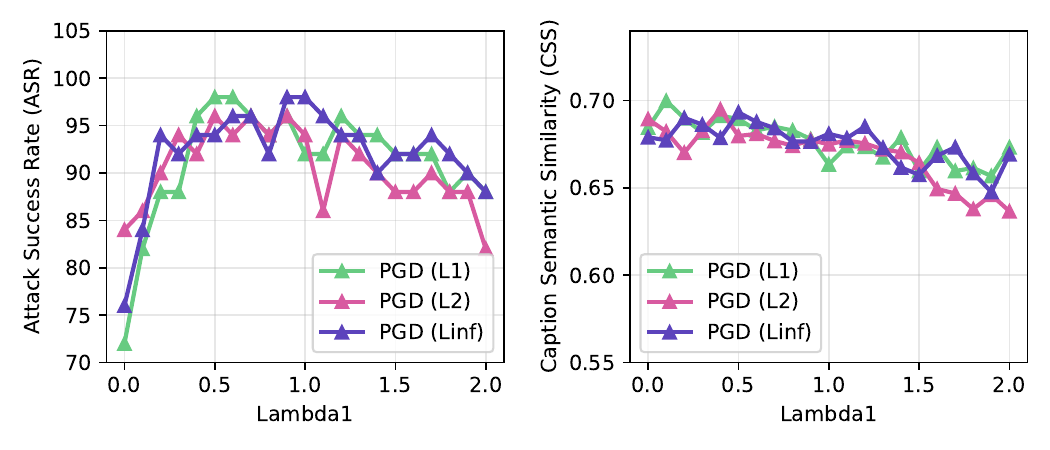}
        \caption{HiPS-cls (Class Labels)}
        \label{fig:lambda_classes}
    \end{subfigure}
    \hfill
    \begin{subfigure}[b]{0.49\textwidth}
        \centering
        \includegraphics[width=\textwidth]{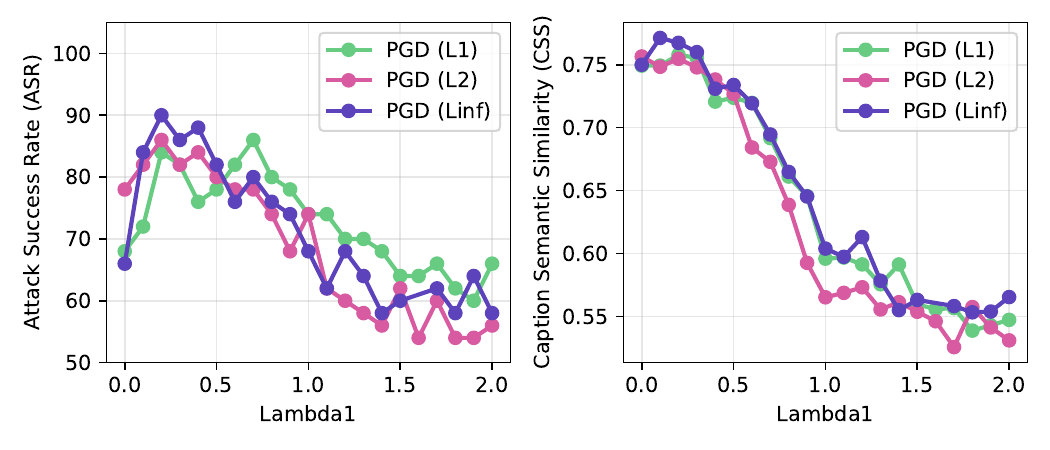}
        \caption{HiPS-cap (Adversarial Captions)}
        \label{fig:lambda_captions}
    \end{subfigure}
    
    \caption{Comparing the sensitivity of hyperparameter $\lambda_1$ on HiPS-cls and HiPS-cap attacks.}
    \label{fig:lambda_sensitivity}
\end{figure}

\textbf{Sensitivity to Lambda:} Figure \ref{fig:lambda_sensitivity} illustrates the effect of the hyperparameter $\lambda_1$ on the performance of different HiPS attacks while keeping $\lambda_2$ fixed at 1.0 for different HiPS PGD variants. We observe that as the magnitude of $\lambda_1$ increases in HiPS-cls, both the ASR and CSS remain relatively high (greater than 90\%) across a wide range of $\lambda_1$ values, from 0.3 to 1.8, across all PGD variants. This stability is due to the increased emphasis on removing the target object as $\lambda_1$ increases. In contrast, for HiPS-cap, increasing $\lambda_1$ places greater focus on reducing the score of the original caption while maintaining the weight of the adversarial caption. As a result, when $\lambda_1$ exceeds 0.5, both ASR and CSS decline rapidly. See Appendix Figure \ref{fig:lambda_sensitivity_torr} and \ref{fig:lambda_sensitivity_iq} for TORR, RORR and image quality metrics.


\section{Limitations, Discussion and Conclusion}

In this work, we demonstrate promising results for the HiPS attack, showing that it is possible to generate small perturbations which cause subtle differences in the output of a downstream task. However, we recognize a few current limitations and future work needed to overcome them. First, the metrics we use to measure success are not always correct. For example, the rule based metrics (TORR and RORR) are biased towards only detecting the presence and absence of an object(s) from a text caption, and does not consider if the sentence is grammatically correct or if additional objects were added to the caption even though they do no exist in the image. In our small dataset this occurs infrequently but can skew the results more on a large dataset. In addition, we find that the cosine similarity metric is not precise enough to measure the small differences between cosine similarities of our caption since they are all very close to each other (by design). In the future, we plan on using a LMM to evaluate the results in a more accurate manner using custom prompts. In addition, in this work we our experiment was restricted to 50 images due to the manual annotations required for generating two adversarial captions for each image. We plan to conduct much larger experiments in the future by automating this process using LMMs, prompting the model to generate a caption with the target object missing. Finally, while in this work we focused on a single image captioning model, we believe that this attack can be used for other multimodal models (LLaVa, OpenFlamingo) as well as other downstream tasks (object detection, action recognition). The transferability of the attack can be improved using an ensemble of multimodal models to generate the attack. 


\section*{Acknowledgments}
This manuscript has been authored by UT-Battelle, LLC, under contract DE-AC05-00OR22725 with the US Department of Energy (DOE). The US government retains and the publisher, by accepting the article for publication, acknowledges that the US government retains a nonexclusive, paid-up, irrevocable, worldwide license to publish or reproduce the published form of this manuscript, or allow others to do so, for US government purposes. DOE will provide public access to these results of federally sponsored research in accordance with the DOE Public Access Plan (
https://www.energy.gov/doe-public-access-plan).

\bibliography{main}
\bibliographystyle{unsrtnat}

\clearpage
\par\vfill\par
\newpage

\appendix

\section{Additional Details of Evaluation Metrics}
Attack Success Rate (ASR) metric is an aggregated evaluation of the success of modifying textual captions to remove references to a target object while preserving mentions of remaining objects.
The aggregation constitutes of two measures: (1) Target Object Removal Rate (TORR) and (2) Remaining Objects Retention Rate (RORR). TORR assesses whether references to a specific target object $T_{\text{target}}$ are effectively removed from the caption generated after the HiPS attack $\tilde{C}$, measuring how well the perturbations obscure $T_{\text{target}}$ from the model's point-of-view.
RORR, on the other hand, evaluates whether references to the remaining objects $T_i$ are preserved in $\tilde{C}$. This ensures that the perturbation does not inadvertently affect or remove references to the objects other than the targeted one.

We calculate the TORR and RORR metrics in two steps: (i) word segmentation and cleaning, (ii) semantic presence validation.

(i) Word segmentation and cleaning: To identify object references within $\tilde{C}$, we tokenize words using $spaCy$ \citep{honnibal2020spacy}, yielding a list of words $W_{\tilde{C}}$. We assume that specific words in $W_{\tilde{C}}$ correspond to object representations. We remove stop words and punctuation as they do not contribute to our evaluation schema and help streamline the analysis by focusing on meaningful words that are critical to understanding the content of $\tilde{C}$. We convert plurals in $W_{\tilde{C}}$ to singular forms using the $inflect$ engine in Python. This normalization aids in matching terms more effectively, when compared to $T_{\text{target}}$ and $T_i$, during semantic presence validation. Finally, we filter $W_{\tilde{C}}$ based on Part-of-Speech tags using $spaCy$. We exclude determiners (DET) and pronouns (PRON) from our analysis as they do not bear any significance to our analysis.
This extensive processing within this step is integral to transforming $\tilde{C}$ into a refined set of lexical units that accurately represent its meaningful content. It allows the ASR metric to perform precise evaluations of object presence, enhancing the accuracy and validity of the analysis and reducing the risk of misinterpretations due to irrelevant or misleading text components.

(ii) Semantic presence validation: Provided $W_{\tilde{C}}$ from step (i), we verify the absence of $T_{\text{target}}$ and the presence of $T_i$. This step involves both direct presence checks and similarity-based assessments. For direct presence check, we perform string-matching comparisons between $W_{\tilde{C}}$ and $T_{\text{target}}$, as well as between $W_{\tilde{C}}$ and $T_i$. If the direct presence check does not yield a clear result, we employ cosine similarity between word embeddings to further validate the success of $T_{\text{target}}$ removal and $T_i$ retention. Using an empirically established similarity threshold (0.7 in this case), we determine the boundary for distinguishing between successful and unsuccessful removal/retention of the objects.
Our ASR metric offers multiple options for obtaining word embedding, including Word2Vec \citep{church2017word2vec}, GloVe \citep{pennington2014glove}, FastText \citep{bojanowski2017enriching}, and BERT \citep{devlin2018bert}, during semantic presence validation, enhancing its adaptability and effectiveness across various downstream tasks and models. For our specific application, we found GloVe to be the most effective choice.

The ASR metric, while robust for many scenarios, encounters challenges when dealing with multi-word objects, such as "teddy bear." In these cases, the metric may struggle to effectively assess the presence or removal of the entire phrase because it traditionally operates on individual word embeddings. Our workaround for this limitation involves averaging the embeddings for each word within the multi-word phrase.
Furthermore, the metric's reliance on cosine similarity thresholds may not fully account for the nuanced differences between conceptually related but distinct objects and vice-verse. For example, while "hills" and "mountains" are closely related, they are not interchangeable.
Our experiments show that the ASR metric might fail to recognize this subtle distinction, leading to rare but incorrect assessments of object removal success.

\section{Hyperparameter Details}
We have used $\lambda_2=1$, in all of our experiments. Additional details of other hyperparameters are provided in Table \ref{tab:hyper}.

\begin{table}[h]
\centering
\caption{Hyperparameter Details for Best Performing Models from Table \ref{tab:asr}}
\label{tab:hyper}
\begin{tabular}{ccccc}
\hline
\multicolumn{1}{l}{} & \multicolumn{2}{c}{HiPS-cls (Class Labels)} & \multicolumn{2}{c}{HiPS-cap (Adv. Caption)} \\ \hline
\multicolumn{1}{l}{} & $\alpha$               & $\epsilon$               & $\alpha$               & $\epsilon$                \\ \hline
FGSM                 & 2/255               & 0.03                  & 2/255               & 0.03                  \\
PGD ($L_1$)          & 500                 & 1000                  & 500                 & 1000                  \\
PGD ($L_2$)          & 5                   & 5                     & 5                   & 10                    \\
PGD ($L_\infty$)     & 2/255               & 0.02                  & 2/255               & 0.06                  \\ \hline
\end{tabular}
\end{table}

\section{Additional Results}

We present some additional results on image quality, TORR, and RORR for the effect of attack budget and hyperparameter sensititivity.

\begin{figure}[h]
\centering
    \includegraphics[width=\linewidth]{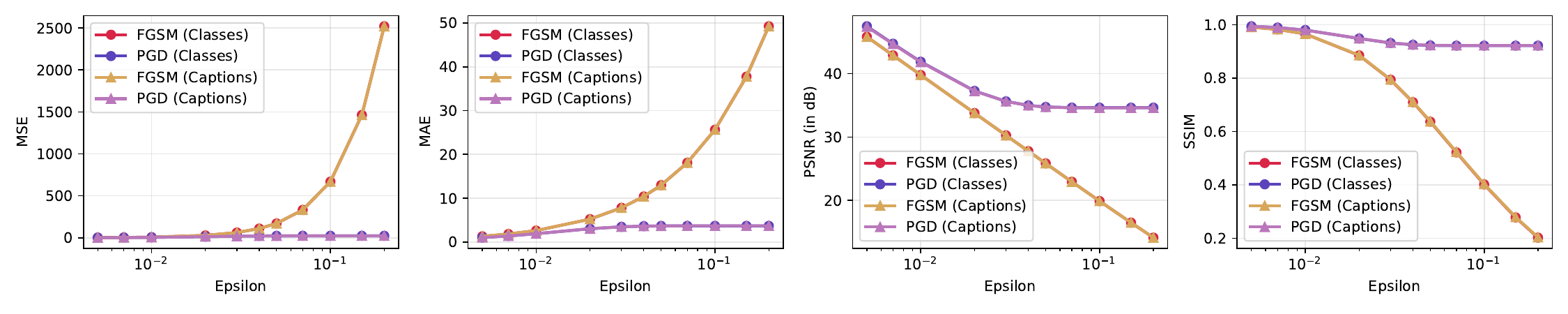}
    \caption{Comparing the effect of attack budget $\epsilon$ on the different image quality metrics for HiPS-cls and HiPS-cap attacks using FGSM and PGD with $L_\infty$ norm.}
    \label{fig:epsilon_sweep_img_qual}
\end{figure}

\begin{figure}[h]
    \centering
    \begin{subfigure}[b]{\textwidth}
        \centering
        \includegraphics[width=\textwidth]{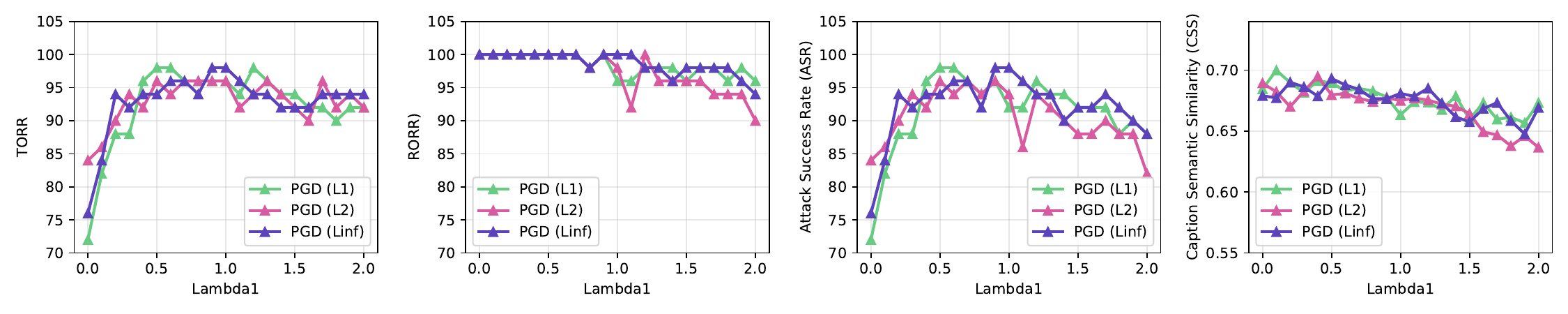}
        \caption{HiPS-cls (Class Labels)}
    \end{subfigure}
    \hfill
    \begin{subfigure}[b]{\textwidth}
        \centering
        \includegraphics[width=\textwidth]{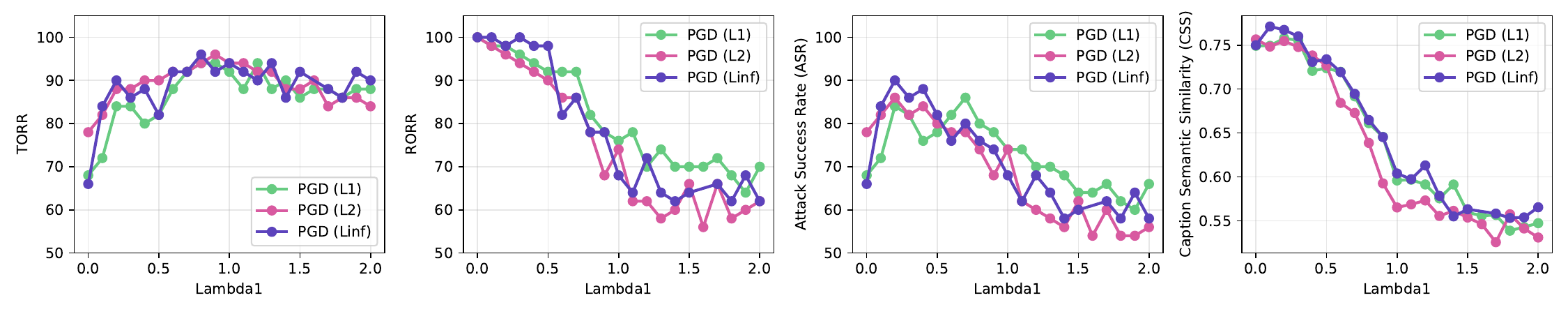}
        \caption{HiPS-cap (Adversarial Captions)}
    \end{subfigure}
    
    \caption{Comparing the sensitivity of hyperparameter $\lambda_1$ on HiPS-cls and HiPS-cap attacks.}
    \label{fig:lambda_sensitivity_torr}
\end{figure}

\begin{figure}[h]
    \centering
    \begin{subfigure}[b]{\textwidth}
        \centering
        \includegraphics[width=\textwidth]{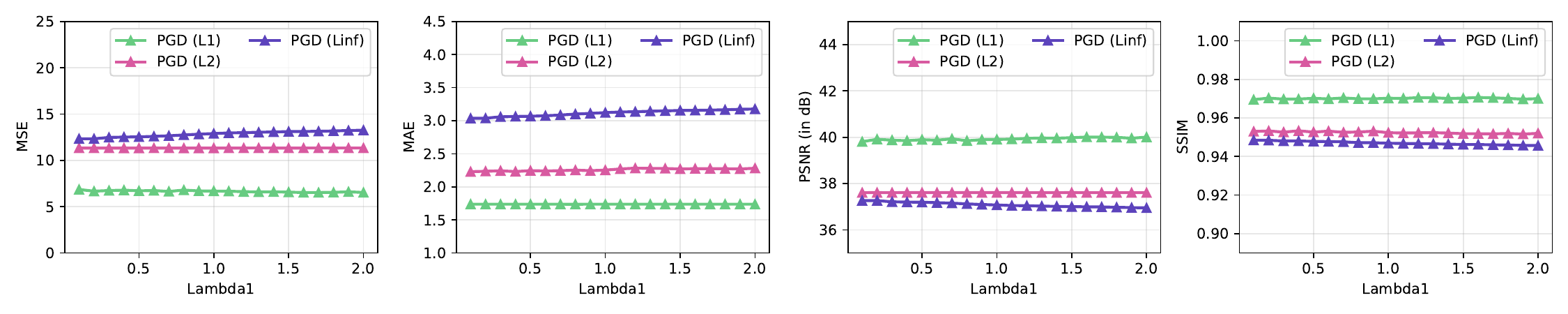}
        \caption{Classes}
        \label{fig:lambda_classes_iq}
    \end{subfigure}
    \hfill
    \begin{subfigure}[b]{\textwidth}
        \centering
        \includegraphics[width=\textwidth]{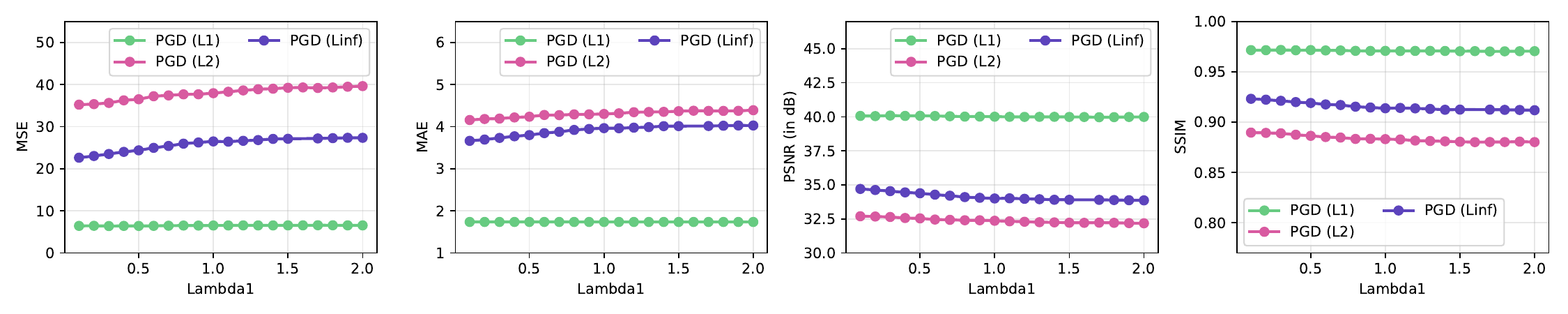}
        \caption{Adversarial Captions}
        \label{fig:lambda_captions_iq}
    \end{subfigure}
    
    \caption{Comparing image quality metrics on the sensitivity of hyperparameter $\lambda_1$ on HiPS-cls and HiPS-cap attacks.}
    \label{fig:lambda_sensitivity_iq}
\end{figure}

\end{document}